\RequirePackage{amsthm}
\documentclass{WileyMSP-template}

\usepackage{graphicx}%
\usepackage{multirow}%
\usepackage{amsmath,amssymb,amsfonts}%
\usepackage{amsthm}%
\usepackage{mathrsfs}%
\usepackage[title]{appendix}%
\usepackage{xcolor}%
\usepackage{textcomp}%
\usepackage{manyfoot}%
\usepackage{booktabs}%
\usepackage{hyperref}
\usepackage{algorithm}%
\usepackage{algorithmicx}%
\usepackage{algpseudocode}%
\usepackage{listings}%
\usepackage{adjustbox}%

\begin{document}

\pagestyle{fancy}

\title{A novel parameter estimation method for pneumatic soft hand control applying logarithmic decrement for pseudo rigid body modeling}

\maketitle


\author{Haiyun Zhang*}
\author{Kelvin HoLam Heung*}
\author{Gabrielle J. Naquila}
\author{Ashwin Hingwe}
\author{Ashish D. Deshpande}


\dedication{The first two authors, Haiyun Zhang and Kelvin HoLam Heung, contributed equally to this work.}

\begin{affiliations}
Haiyun Zhang, Gabrielle J. Naquila, Ashwin Hingwe, Ashish D. Deshpande\\
Department of Mechanical Engineering, The University of Texas at Austin, 2515 Speedway, Austin, TX 78712, United States\\
Email Address: haiyunzhang@utexas.edu; ashish@austin.utexas.edu

Kelvin HoLam Heung\\
Department of Building and Real Estate, the Hong Kong Polytechnic University, 11 Yuk Choi Roa, Hung Hom, Hong Kong, 10587, China
Email Address: kelvin.heung@polyu.edu.hk
\end{affiliations}



\begin{abstract}

The rapid advancement in physical human-robot interaction (HRI) has propelled the growth of soft robot design and, in parallel, soft robot controllers. Controlling soft robots, especially soft hand grasping, is complex due to their ubiquitous deformation, prompting the use of reduced model-based controllers to provide sufficient state information for real-time high dynamic response control performance. However, most modeling techniques face computational efficiency and complexity of parameter identification issues and are hard to apply to real-time controls. To alleviate this, we proposed a paradigm coupling an analytical modeling approach based on Pseudo-Rigid Body Modeling and the Logarithmic Decrement Method for parameter estimation (PRBM+LDM). Using a soft robot hand test bed, we demonstrate the accuracy of PRBM+LDM to model position and force output as a function of pressure input and benchmark its performance. We, then, apply the PRBM+LDM model as a basis for a closed-loop position controller and compare its performance against a simple PID controller. Furthermore, we apply the PRBM+LDM model as a closed-loop force controller and compare its performance with simple constant pressure grasping control by performing small contact areas pinching tasks on low-weight, small objects - a screwdriver, a potato chip, and a brass coin. The PRBM+LDM-based position controller (Average Max. Error across all fingers: $4.37\degree$) outperformed the simple PID position controller (Average Max. Error across all fingers: $20.38\degree$). Furthermore, the PRBM+LDM-based force controller (Potato chip: 86\%, Screwdriver: 74.42\%, Brass coin: 64.75\%) achieved a higher success rate than the constant pressure grasping control  (Potato chip: 82.5\%, Screwdriver: 70\%, Brass coin: 35\%) in the pinching tasks. We conclude that the PRBM+LDM modeling technique proves to be a convenient and efficient way to model the dynamic behavior of soft actuators closely and can be used to build high-precision position and force controllers. In application, it realizes stable, flexible grasping of small objects by exerting precise contact force on contact areas. 

\noindent\textbf{Published version:} This is the author-accepted manuscript of the article published in \textit{Advanced Intelligent Systems}. 
The final version is available at Wiley Online Library: \href{https://doi.org/10.1002/aisy.202400637}{https://doi.org/10.1002/aisy.202400637}.
\end{abstract}


\section{Introduction}\label{sec1}

The surge in physical human-robot interaction (HRI) has given rise to the necessity of compliant robotic designs to ensure safety in different human-robot workspaces \cite{Das2019}. This demand for compliant systems has catalyzed a wave of soft robots that provide inherent compliance and flexibility. Soft robots have been ubiquitously exploited in the last decade, and their applications have penetrated numerous robotics fields, such as surgical \cite{Kim2013, Low2014, Liu2016, Runciman2019}, industrial \cite{Schmitt2018}, and assistive robotics \cite{Pan2022}, where HRI is prevalent and necessary. As the sophistication of soft robot design continues, works on soft robot controllers have also been vigorously explored in parallel. However, controlling soft robots presents unique challenges compared to rigid bodies, as soft bodies exhibit complex deformations beyond simple planar motions. Unlike rigid bodies, which can be described by six degrees of freedom (three rotations and three translations about the x, y, and z axes), soft materials are elastic and can bend, twist, stretch, compress, buckle, wrinkle, and more. This variability in motion offers an infinite number of degrees of freedom, demanding innovative approaches to modeling, control strategies, dynamics analysis, and high-level planning \cite{Rus2015}.

To attack the complexity of soft robot control, scientists have been exploring model-based and model-free controllers. Model-based controllers depend on analytical models for deriving the controller, while model-free controllers circumvent the need to use kinematic and dynamic models \cite{Thurutel2018}. Historically, model-free controllers predominantly used open-loop valve sequencing to control body-segment actuation. Valve sequencing means that a valve is turned on for some duration of time to pressurize the actuator and then turned off to either hold or deflate it\cite{Thurutel2018}. This type of control has also been referred to as bang-bang control\cite{Shepherd2011,onal2017soft,martinez2013robotic,tolley2014resilient,correll2014soft}. In recent years, model-free controllers have been focusing on using learning-based techniques\cite{Baghat2019, Iyengar2020, Morimoto2021, Zhao2021, Li2022, AlAttar2023, Almanzor2023}. However, for most applications, model-based controls have been applicable since simplified models provide enough information to improve control performance significantly compared to the model-free baseline. Furthermore, many finite-dimensional modeling techniques have proven to be accurate, manageable, and interpretable \cite{Santina2023} to be used over model-free controllers. 

Model-based static controllers are currently the most widely used and studied strategy for the control of soft robots \cite{Thurutel2018}. Analytical models using Constant Curvature (CC) modeling \cite{Trivedi2008, Robert2010, Camarillo2009, Kapadia2011, Kapadia2014, Penning2011, Penning2012}, Piecewise Constant Curvature (PCC) modeling \cite{Hannan2003, Li2018, Jones2006}, Cosserot rod theory modeling \cite{Trivedi2008, Renda2018, Xu2022, Roshanfar2023}, disc-thread modeling \cite{Bui2021, Schultz2022}, and 3D Finite Element Models \cite{Connolly2015, Duriez2013, Vavourakis2011} have been vastly explored. However, using Cosserot rod theory and disc-thread modeling presents an overall increased computational difficulty. On the other hand, Finite Element Analysis (FEA) simulations are time-consuming, and parameter identification is dependent on tensile testing the individual materials of the actuator separately, which is also time-consuming and leads to an increase in parameter deviation from a fully assembled actuator. Furthermore, FEA assumes a quasistatic analysis which leads to further deviation from the actual behavior of the soft actuator. Across the literature, CC and PCC modeling are the most prevalent methods applied in model-based controllers due to their relative simplicity \cite{Thurutel2018}. However, their non-linear analysis introduces complexity that may not be justifiable for certain applications, similar to the Cosserot rod theory and disc-thread modeling techniques. Furthermore, CC and PCC only apply when external loading effects are negligible, further constraining their applicability.

For physical HRI applications, it is imperative to have a controller that can be applied conveniently, comparable to model-free methods, and can provide sufficient baseline accuracy, much like model-based controllers. Therefore, the computational efficiency of the model, the convenience of applying the modeling technique, and the ability to properly model the dynamics of the soft actuator under considerable external load and under a non-quasistatic state are crucial for effective control system development. Hence, in this work, we employ the Pseudo-Rigid Body Modeling (PRBM), introduced in \cite{Bandopadhya2008} and \cite{Tang2019}, to build the dynamic equation for model-based control of soft robots and complement it with the Logarithmic Decrement (LDM) technique for parameter estimation. PRBM is a common method used to account for large deformations in compliant mechanisms. By approximating a compliant mechanism with rigid links and torsional springs and outlining a linear model, PRBM simplifies the analysis of a compliant mechanism by an appropriate equivalent rigid-body mechanism \cite{Howell2013}. We then leverage the LDM technique to help us obtain the stiffness and damping parameters for the dynamic model acquired from PRBM. LDM quantifies the rate of reduction of the oscillating amplitude of a system \cite{Jadhav2019}, providing a measure of the system's damping characteristics. Onwards, this manuscript will refer to the coupled method of Pseudo-Rigid Body Modeling and Logarithmic Decrement as the PRBM+LDM method.
\begin{figure}[htbp]%
\centering
\includegraphics[width=1.0\textwidth]{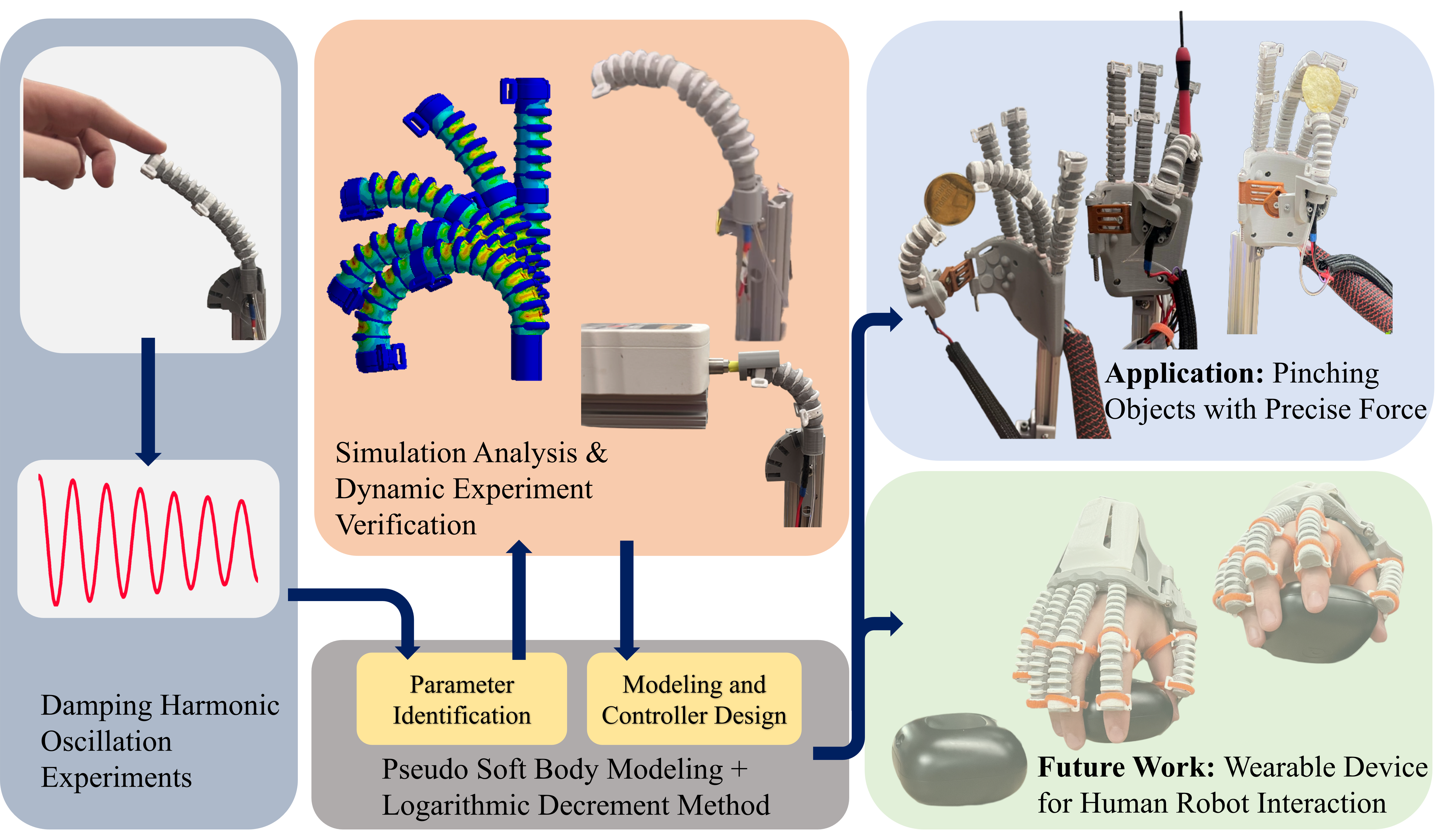}
\caption{Overview of the research workflow from data acquisition, and construction of Pseudo Rigid Body Modeling and Controller design, to the application and future work.}
\label{Fig1}
\end{figure}

To demonstrate the effectiveness and practicality of the PRBM+LDM method in dynamic modeling soft actuators, we utilize a customized pneumatic soft robot hand test bed. Through a series of experiments conducted on the soft robot hand, we validate the ability of the PRBM+LDM to model positional and force dynamics accurately. Subsequently, we apply the PRBM+LDM as a closed-loop position controller and a closed-loop force controller. To benchmark the position controller, we compare its position accuracy with that of a simple PID controller. To verify the functionality of the PRBM+LDM-based force controller, we perform a pinching task with various objects, i.e., a screwdriver, a brass coin, and a potato chip (Supplementary Video 1), and compare its performance with a simple constant pressure grasping control (Fig. \ref{Fig1}).

\section{Results}\label{sec2}
\subsection{Dynamic Equation and Stiffness and Damping Parameter Validation}\label{subsec2.1}
\subsubsection{Stiffness and Damping Coefficient Acquisition}\label{subsubsec2.1.1}

To apply the PRBM+LDM method, our study utilizes a soft robot test bed to develop our analytical model. The test bed features a soft robotic hand, which includes five beam-like fingers made from silicone. These fingers are designed for pneumatic actuation, with each finger containing a hollow chamber to accommodate variable air pressure input. A detailed description of the soft robotic hand's design is elaborated in the Methods section Mechanical Design of the Soft Robot Hand Test Bed.

\begin{figure}[htbp]%
\centering
\includegraphics[width=0.95\textwidth]{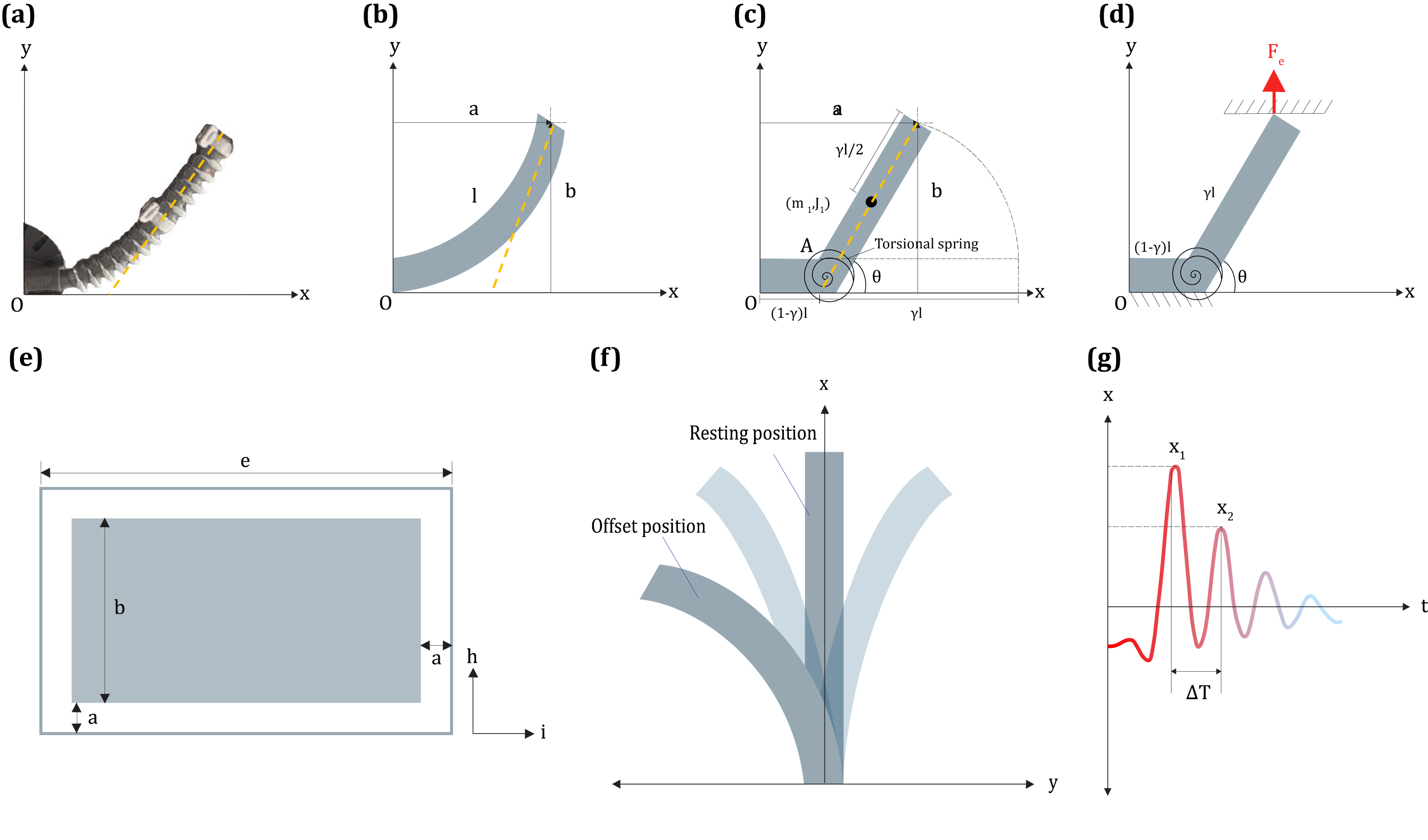}
\caption{Pseudo Rigid Body Modeling of a soft robot hand finger. \textbf{(a)} Index soft finger of the soft robot hand test bed, \textbf{(b)} compliant body diagram of a beam-shaped soft finger, \textbf{(c)} PRBM equivalent rigid body model of the compliant beam-shaped soft finger, \textbf{(d)} Force exertion by the soft finger to the environment, \textbf{(e)} Cross-sectional diagram of the soft finger, \textbf{(f)} Offsetting fingertip of finger actuator from rest position for LDM experiment, \textbf{(g)} Underdamped harmonic oscillation of a soft finger.}
\label{fig2}
\end{figure}

To design the analytical model, we use the PRBM on each soft finger. Fig. \ref{fig2}a-d shows a representation of the PRBM diagram of a finger. Concurrently, Equation \ref{eq:transferfunction} denotes the transfer function of the finger’s output bending angle as a function of the input pressure as derived in the Methods section Derivation of Dynamic Equation through Pseudo-Rigid Body Modeling. 

\begin{equation}
    \label{eq:transferfunction}
    \frac{\theta(s)}{P_{in}(s)}=\frac{\frac{N}{A}}{s^2+D s+\frac{K}{A}}
\end{equation}

$P_{i n}$ is the input pressure in the chamber of the finger, $\theta$ is the bending angle of the finger, $N=\int_{a+l_{a r m}}^{a+b+l_{a r m}}(e-2 a) \cdot h \cdot d h$, $A=\frac{7}{12}m \gamma^3 l^2$ is a constant value, $\gamma$ is the length ratio of the finger, the mass of the finger $m, l$ is the length of the finger, $D$ is the damping factor, and $K$ is the stiffness value.

The soft finger’s transfer function obtained through PRBM is a second-order transfer function, from which we can use its characteristic equation to obtain the damping factor and the stiffness coefficient. The characteristic equation is analogous to the conventional second-order characteristic equation, $s^2+2 \varepsilon \omega_n s+\omega_n^2=0$, where $\omega_n$ is the natural frequency, $\omega_d$ is the damped frequency, and $\varepsilon$ is the damping ratio. The LDM helps us obtain the stiffness and damping parameters of the second-order characteristic equation. By definition, the LDM $\delta=\frac{1}{n} \ln \left(\frac{x_1}{x_2}\right)$ where $x_1$ and $x_2$ are the amplitudes of two consecutive peaks of the underdamped oscillation, and $n$ represents the number of oscillations between these peaks. We can relate the LDM with the $\omega_n$ through the damping ratio $\varepsilon=\frac{\delta}{\sqrt{4 \pi^2+\delta^2}}$,   and then use the value of $\varepsilon$ to find $\omega_n=\frac{\omega_d}{\sqrt{1-\varepsilon^2}}$, where $\omega_d=\frac{2 \pi}{T}$. Since $\omega_n^2=\frac{K}{A}$, we finally obtain the stiffness coefficient, $K$. This leads us to find the damping coefficient, as well, since $D=2\varepsilon\omega_n$. Reviewing the equations, we simply need to obtain $x_1$, $x_2$, and $\Delta T$ experimentally through the harmonic response of the soft finger to determine the stiffness and damping parameters for the dynamic equation of the soft finger.

\begin{table}[bt]
\caption{Average estimated stiffness and damping coefficients of the five soft robot hand fingers obtained using LDM }
\begin{tabular}{|p{1.0cm}|p{3.5cm}|p{3.5cm}|}
\hline
Name& Stiffness coefficient, $k\_est$ (Nm/rad)& Daming coefficient, $d\_est$ (Nm.s/rad)\\ \hline
Thumb  & 0.6541 & 0.0011    \\ \hline
Index  & 0.570  & 0.0031    \\ \hline
Middle & 0.5778 & 0.00127   \\ \hline
Ring   & 0.5069 & 0.00178   \\ \hline
Little & 0.5886 & 0.0011224 \\ \hline
\end{tabular}
\label{tab1}
\end{table}

To acquire the harmonic response of the soft finger, we push the tip to offset the soft finger from its resting position and release it to capture its harmonic oscillatory behavior (Fig. \ref{fig2}f-g). The details of the experimental setup are described in the Methods section Experimental Setup of Logarithmic Decrement Technique. Capturing each soft finger’s harmonic oscillation behavior for 10 trials and finding their average stiffness and damping factors, we obtain the stiffness and damping coefficients in Table \ref{tab1}, including the geometric and harmonic oscillation characteristics of each soft finger. Supplementary Table 1 extends the content of Table 1, including the geometric and harmonic oscillation characteristics of each soft finger.

\begin{figure}[htbp]%
\centering
\includegraphics[width=0.95\textwidth]{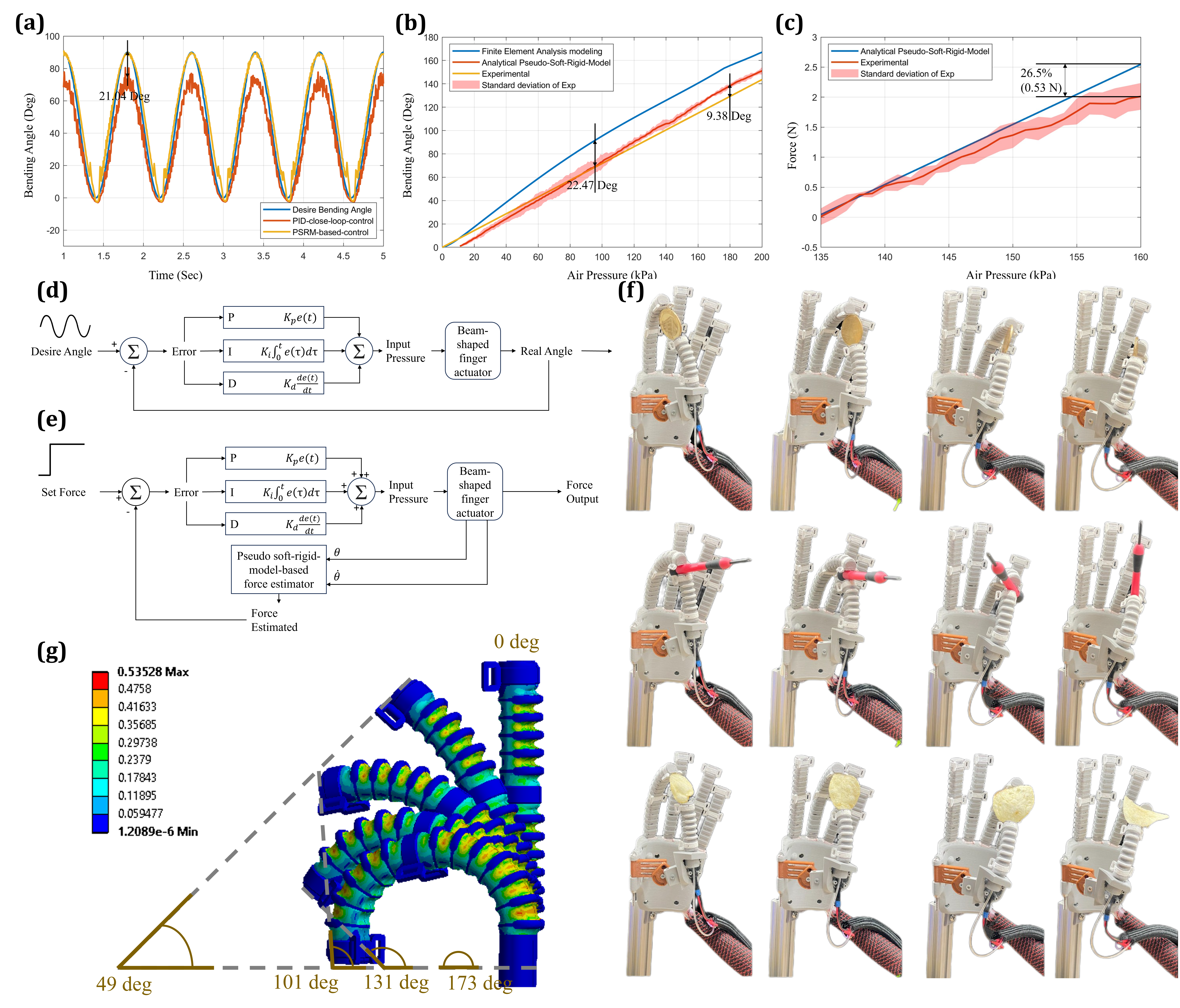}
\caption{(a)\textbf{ }Index finger angular position tracking of closed-loop PRBM+LDM-based controller compared with a simple PID controller’s angular position tracking, \textbf{(b) }PRBM+LDM modeling of the index finger’s angular position as a function of pressure input compared with FEA model and experimental angular position values,\textbf{(c) }PRBM+LDM modeling of the index finger’s output force as a function of pressure input compared with experimental force values, \textbf{(d) }PRBM+LDM-based closed-loop position controller\textbf{, (e) }PRBM+LDM-based closed-loop force controller, \textbf{(f) }Demonstration of the soft robot hand performing pinching tasks with a brass coin, screwdriver, and potato chip using the PRBM+LDM-based force controller. \textbf{(g)} FEA simulation result shows the pressure vs bending angle relationship of index and ring fingers which apply the same beam-shaped finger configuration.  The pressure applied to the thumb ranges from 0 to 200 kPa. The five FEA images are captured at moments evenly distributed among the range of pressure. }
\label{fig3}
\end{figure}
\subsubsection{Stiffness and Damping Coefficient Benchmarking}
To validate the stiffness and damping coefficients obtained from the PRBM+LDM method, we compare the theoretical position output of each soft finger's dynamic equation with experimental data and FEA, as well as the theoretical force output with experimental results. To execute the benchmarking experiments, we analyze the position and force output based on the equation of motion of the soft fingers as shown in Equation \ref{eq2}, where $M$ is the moment generated by air pressure, $J=\left[\begin{array}{ll}-\frac{\gamma l}{2} \sin \theta_1 & -\frac{\gamma l}{2} \cos \theta_1\end{array}\right]$ is the Jacobian matrix mapping the angular velocity $\dot{\theta}$ to the end-effector's velocity $\left[\begin{array}{c}\dot{x}_p \\ \dot{y}_p\end{array}\right]$, and $F_e=\left[\begin{array}{ll}F_x & F_y\end{array}\right]^{\top}$ is the force exerted by the soft finger.

\begin{equation}
    A \ddot{\theta}+B \dot{\theta}+K \dot{\theta}=M-J^{\top} F_e
\label{eq2}
\end{equation}

Since the solenoid valves connected to the finger actuators have a slow response rate $(10 \mathrm{~Hz})$, we can assume a quasi-static behavior with the angular velocity and angular acceleration are approximately $\ddot{\theta}=\dot{\theta}=0$.

\begin{equation}
P_{\text {des }}=\frac{A K \theta}{N}
\label{eq3}
\end{equation}

For position modeling, we assume there is no external force resisting the finger actuator. Therefore, we can set $F_e=0$. With $M=\frac{N}{A} P_{i n}$, where $P_{i n}$ is the input pressure, Equation \ref{eq2} simplifies to $K \theta=\frac{N}{A} P_{i n}$ where we can map the angular position, $\theta$, with the input pressure as expressed in Equation \ref{eq3}.

In force modeling, the soft finger's tip experiences a non-zero force due to the reaction forces encountered upon contacting an object. Therefore, the relationship of the soft finger's force output, $F_e$, with the pressure input can be expressed as seen in Equation \ref{eq4}.

\begin{equation}
P_{d e s}=\frac{A\left(K \theta+J^T F_\epsilon\right)}{N}
\label{eq4}
\end{equation}

The experimental setups to benchmark the theoretical output values of the position and force modeling are detailed in the Methods section Position and Force Output Modeling Experimental Setup, while the FEA configuration to benchmark the open-loop position control is outlined in the Methods section Finite Element Analysis of Soft Fingers.  

The theoretical position output values we obtain using the PRBM+LDM method closely match the experimental values for all five soft fingers (Fig. \ref{fig3}b and Supplementary Table \ref{tab2}), with Root Mean Square Errors (RMSE) of $4.78\degree$ and a maximum error of $8.82\degree$ across all fingers. Conversely, the FEA values exhibit a larger deviation from the experimental values, with an average RMSE of $14.78\degree$ and a maximum error average of $22.73\degree$ .  

On the other hand, the values of the theoretical force output values we obtained using the PRBM+LDM method exhibit an average RMSE of 0.324 N and an average maximum error of 0.58 N from the experimental force output values (Fig. \ref{fig3}. c and Supplementary Table 3).  

\subsection{PRBM+LDM Position Control Application }

Having benchmarked the stiffness and damping coefficients obtained using the PRBM+LDM method and demonstrated its ability to model position and force output as a function of pressure input, we now explore its feasibility as a position controller for the soft robot hand. In Fig. \ref{fig3}d, we use the PRBM+LDM as a feedforward controller with a feedback PID loop design with the assumption that ${\dot{\theta}}=0$ in order to model a non-quasistatic state. While the PRBM+LDM feedforward model converts the desired angular position of the soft finger to a pressure input, the PID loop is added to compensate for the error from the forward model. The PID is designed as a backward close-loop and its output is summed to the output of the feedforward PRBM+LDM model. We compare the PRBM+LDM-based position controller design with a bare single-loop PID controller tuned by using the auto-tuning tool of MATLAB Simulink 2020R and further manual tuning. With a sinewave reference trajectory that ranges from $0\degree$ to $90\degree$ with a period of 0.75s, we apply the position controllers to each of the soft robot hand’s fingers with the same experimental setup as described in the Methods section Position and Force Output Modeling Experimental Setup. In Fig. \ref{fig3}a, we can see the position modeling accuracy of both the PRBM+LDM-based controller and the bare PID controller compared with the reference angular position trajectory for the index finger of the soft robot hand. The PRBM+LDM-based controller tracks the reference trajectory (Table \ref{tab2}), with an average RMSE of $2.37\degree$ and an average maximum error of $4.37\degree$, closer than the bare PID controller, with an average RMSE of $16.51\degree$ and an average maximum error of $20.38\degree$.

\begin{table}[ht]
 \centering
\caption{Comparison of the PRBM+LDM-based position controller performance versus the PID-based position controller}
   
\begin{tabular}{|p{1.0cm}|p{2.5cm}|p{2.5cm}|p{2.5cm}|p{2.5cm}|}
\hline Finger &  
Max. Error for 
PRBM+LDM 
control (deg)
&  RMSE for 
PRBM+LDM 
control (deg)& 
Max. Error for 
PID control 
(deg)
& 
RMSE for PID 
control (deg)\\
\hline Thumb & 2.36 & 0.92 & 15.51 & 12.23 \\
\hline Index & 4.52 & 3.26 & 21.04 & 18.65 \\
\hline Middle & 7.35 & 4.12 & 24.56 & 19.71 \\
\hline Ring & 3.96 & 2.31 & 22.56 & 17.32 \\
\hline Little & 3.65 & 1.23 & 18.23 & 14.53 \\
\hline
\end{tabular}
\label{tab2}
\end{table}

 \subsection{PRBM+LDM Force Control Application }
 Complementing the position controller, we now further apply the PRBM+LDM method as a closed-loop force controller to the soft robotic hand. Force control is challenging to realize in model-based controllers for soft robots due to the difficulty in fabricating appropriate force sensors on the surface of the actuators. To alleviate this, we design a force controller (Fig. \ref{fig3}e) where we establish a force estimator to approximate the force exerted at the soft finger's tip using the pressure input $P_{i n}$ and measured angular position $\theta_{\text {real }}$ and velocity $\dot{\theta}_{\text{real}}$. The force estimated is expressed in Equation \ref{eq4} such that the $\left(J^T\right)^{+}$is the Moore-Penrose inverse of $J^T$.
 \begin{equation}
F_{\text {est }}=\left(J^T\right)^{+}\left(M-K \theta_{\text {real }}\right)
\label{eq5}
\end{equation}
We experimentally inspect the performance of the force controllers by applying them to the soft robotic hand to perform pinching tests on a screwdriver, a brass coin, and a potato chip (Supplementary Videos 2-4). Handling low-weight and small objects is commonly hard to achieve using soft actuators due to its need for precise force sensing. Soft robots ubiquitously have separate actuation and sensing elements \cite{Polygerinos2017}. This is a result of the compliance of the soft systems, which complicates integration and reduces the reliability of the sensors that need to be embedded in the soft actuator, therefore placing considerable constraints on the design of both the sensors and actuators \cite{zou2024}. We perform a pinching test on a screwdriver to test the soft robot hand’s ability to handle objects with their center of mass away from the contact area. Additionally, we perform a pinching test on a potato chip to test the ability of the soft robot hand to output a precise small-scale force that is large enough to constrain the object but small enough to avoid damaging it. Lastly, the pinching test on the brass coin is to demonstrate if the soft robot hand can generate forces large enough to counteract gravity, ensuring the coin remains constrained at its thin edge without slipping. Around ten successive trials are executed for each object (Fig. \ref{fig3}f) to test the repeatability of the force controller, defining success as being able to pinch the object for at least 5s seconds. The PRBM+LDM-based closed-loop force controller has a comparable success rate with the simple constant pressure control for pinching the potato chip and the screwdriver, and has a better performance for pinching the brass coin. The PRBM+LDM has an 86\% success rate in pinching the potato chip, 74.42\% in the screwdriver, and 64.75\% in the brass coin (Table \ref{tab3}), while the simple constant pressure control exhibits an 82.5\% success rate in pinching the potato chip, 70\% in the screwdriver, and 35\% in the brass coin (Table \ref{tab4}).

\begin{table}[htbp]
\caption{Results of the PRBM+LDM-based force control pinching test by close-loop force controller}
\label{tab3}
\centering
\resizebox{\textwidth}{!}{%
\begin{tabular}{|c|c|c|c|}
\hline
Finger Pair & Potato Chip & Screwdriver & Brass Coin \\
\hline
Thumb \& Index & 62\% (Success: 8, Total trials: 13) & 66.67\% (Success: 8, Total trials: 12) & 73\% (Success: 8, Total trials: 11) \\
Thumb \& Middle & 82\% (Success: 9, Total trials: 11) & 100\% (Success: 11, Total trials: 11) & 67\% (Success: 8, Total trials: 12) \\
Thumb \& Ring & 100\% (Success: 10, Total trials: 10) & 75\% (Success: 6, Total trials: 8) & 42\% (Success: 5, Total trials: 12) \\
Thumb \& Little & 100\% (Success: 11, Total trials: 11) & 56\% (Success: 5, Total trials: 9) & 77\% (Success: 10, Total trials: 13) \\
\hline
\end{tabular}
}
\end{table}
\begin{table}[htbp]
\caption{Results of the simple constant pressure force control pinching test}
\label{tab4}
\centering
\resizebox{\textwidth}{!}{%
\begin{tabular}{|c|c|c|c|}
\hline
Finger Pair & Potato Chip & Screwdriver & Brass Coin \\
\hline
Thumb \& Index & 70\% (Success: 7, Total trials: 10) & 70\% (Success: 7, Total trials: 10) & 40\% (Success: 4, Total trials: 10) \\
Thumb \& Middle & 80\% (Success: 8, Total trials: 10) & 80\% (Success: 8, Total trials: 10) & 30\% (Success: 3, Total trials: 10) \\
Thumb \& Ring & 80\% (Success: 8, Total trials: 10) & 70\% (Success: 7, Total trials: 10) & 30\% (Success: 3, Total trials: 10) \\
Thumb \& Little & 100\% (Success: 10, Total trials: 10) & 60\% (Success: 6, Total trials: 10) & 40\% (Success: 4, Total trials: 10) \\
\hline
\end{tabular}
}
\end{table}

\section{Methods}

\subsection{Mechanical Design of Soft Robot Hand Test Bed }
\begin{figure*}[htbp]%
\centering
\includegraphics[width=0.95\textwidth]{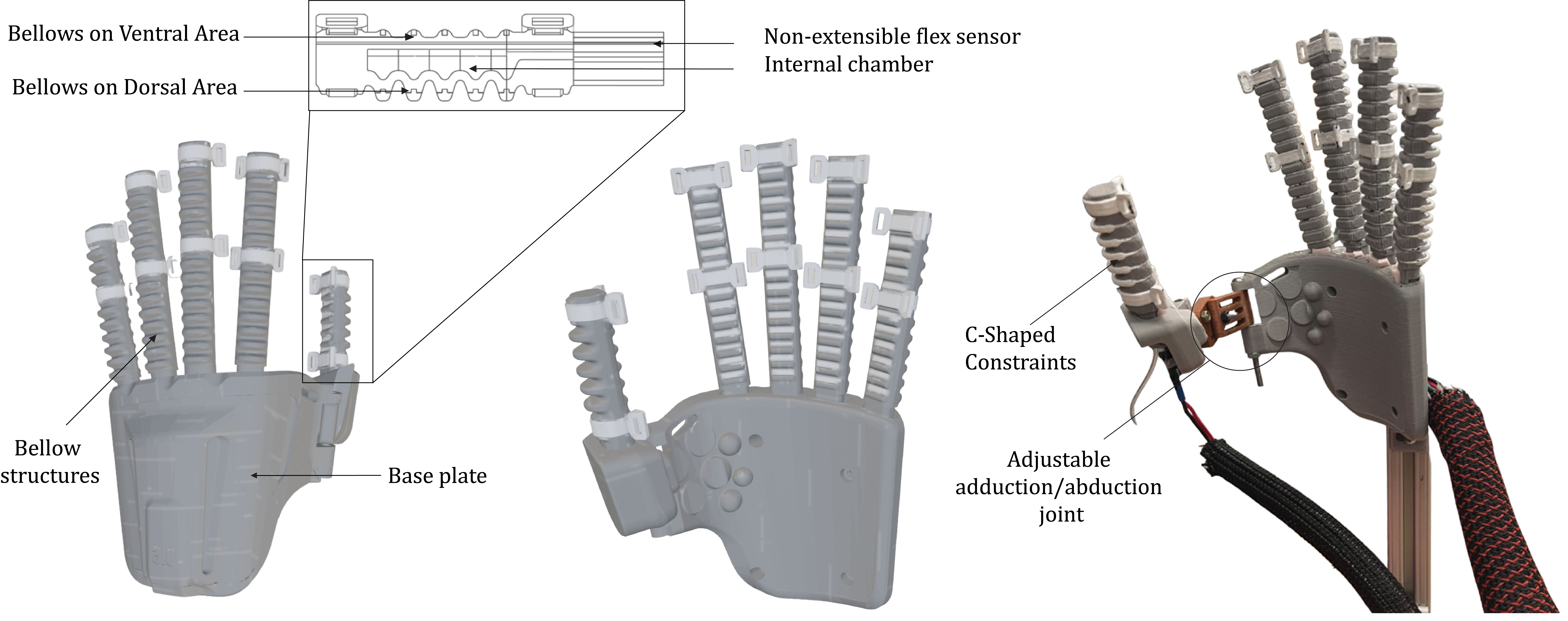}
\caption{Parts of the soft robot hand test bed used in this work, including the cross-sectional view of the thumb representing the internal structure of all the soft fingers. }
\label{fig4}
\end{figure*}

We use a pneumatically driven five-finger soft robotic hand system (Fig. \ref{fig4})\textbf{ }as a test bed for our experiments. The five beam-shaped fingers are made of linearly elastic (SFig. 1) silicone rubber material (ACEO silicone shore A30) with the finger lengths mimicking the proportionality of the lengths of actual human fingers. Each finger has bellow structures on its external layer throughout its circumference and a hollow chamber in its internal layer that allows for the inflow of air pressure. The internal layer’s top surface has bellow structures conforming to the structure of the external layer. However, the bottom surface of the internal layer is flat. To constrain the finger’s radial and axial elongation, ring structures are placed between the bellows of the external layer, and the bottom surface of the internal layer is fixed to a non-extensible flex sensor embedded in a slot between the bottom surface of the internal layer, and the bottom surface of the external layer. The five fingers are grounded on a base plate that mimics the palm of the human hand. The thumb soft finger is grounded to the base plate with an adjustable abduction and adduction motion mechanism, closely mimicking the range of motion of a human thumb. When the internal chamber is pressurized with air, each soft finger bends downward in a piece-wise curvature, influenced by the mechanical constraints of the ring structures and the flex sensor. 

The flex sensor (Flex Sensor - SEN-10264) measures the bending angle of each soft finger. Employing the Wheatstone bridge principle, the flex sensor is connected in series with a 1kOhm resistor to form a voltage divider, with the analog output reflecting the bending deformation. The raw analog voltage output (mV) is then correlated with the corresponding bending angle (deg) through linear regression analysis. This mapping enables the determination of the bending angle of the beam-shaped finger actuator based on the range of raw voltage output. Furthermore, a Butterworth filter is applied to each raw analog signal to eliminate noise artifacts.

In Supplementary Fig. 1, the silicone material closely exhibits a linear stress-strain property (R\textsuperscript{2 }= 0.95921). To reinforce this linearity, the non-extensible flex sensors placed in the beam-shaped fingers also act as a rigid plate with elastic material properties. Therefore, both the silicone material and the rigid plate impose the linearity of the dynamic model. This linearity serves as an important foundation for applying PRBM+LDM method. 
\subsection{Derivation of Dynamic Equation through Pseudo-Rigid Body Modeling }
In this study, we apply the PRBM analytical modeling based on \cite{Tang2019} to obtain the transfer function relationship between the pressure input and angular position of the soft actuator. PRBM is commonly used to analyze compliant mechanisms with linear elastic behavior \cite{hargrove2022}. Corollary, our model assumes a linear relationship between the pressure input and the soft actuator's behavior. PRBM models the bending of a beam-shaped soft actuator as a fixed joint rotation between two connected rigid segments (Fig. 2b-d). The two segments are described by the length ratio, $\gamma$, as $(1-\gamma) l$ and $\gamma l$, where $l$ is the length of the entire soft actuator.

The equation of motion is obtained using the Lagrange equation, $L=K E-P E$, where $K E$ denotes the total kinematic energy and $P E$ denotes the total potential energy.

We obtain the translational $K E$ as follows:
$$
\begin{aligned}
\text { Translational K.E. } & =\frac{1}{2} m\left(\dot{x}^2+\dot{y}^2\right) \\
& =\frac{1}{2} m \gamma\left(\frac{\gamma^2 l^2}{4} \sin ^2 \theta \cdot \dot{\theta}^2+\frac{\gamma^2 l^2}{4} \cos ^2 \theta \cdot \dot{\theta}^2\right) \\
& =\frac{1}{2} m \gamma\left(\frac{\gamma^2 l^2}{4} \cdot \dot{\theta}^2\right) \\
& =\frac{1}{8} m \gamma^3 l^2 \cdot \dot{\theta}^2
\end{aligned}
$$

Where:
$$
\left\{\begin{array}{l}
x=(1-\gamma)+\frac{\gamma l}{2} \cos \theta \\
\dot{x}=-\frac{\gamma l}{2} \sin \theta \cdot \dot{\theta} \\
y=\frac{\gamma l}{2} \sin \theta \\
\dot{y}=\frac{\gamma l}{2} \cos \theta \cdot \dot{\theta}
\end{array}\right.
$$

Since the Total $K E=$ Translational $K E+$ Rotational $K E$

We then obtain the rotational $K E$ as:
$$
\text { Rotational K.E. }=\frac{1}{2} J \dot{\theta}^2=\frac{1}{2} m \gamma\left(\frac{1}{3} m \gamma^3 l^2\right) \dot{\theta}^2=\frac{1}{6} m \gamma^3 l^2 \dot{\theta}^2
$$

Adding the translation and rotational $K E$ together gives us the total $K E$ :
$$
\text { Total K. E. }=\frac{1}{8} m \gamma^3 l^2 \cdot \dot{\theta}^2+\frac{1}{6} m \gamma^3 l^2 \cdot \dot{\theta}^2=\frac{7}{24} m \gamma^3 l^2 \cdot \dot{\theta}^2
$$

The total $P E$ only has a rotational aspect for our purposes. Therefore, we define $P E=\frac{1}{2} k \dot{\theta}^2$.

The Lagrange then is expressed as
$$
L=K . E-P . E .=\frac{7}{24} m \gamma^3 l^2 \cdot \dot{\theta}^2-\frac{1}{2} k \dot{\theta}^2
$$

The governing equation of the soft actuator can then be defined as follows
$$
\begin{gathered}
\frac{d}{d t}\left(\frac{\partial L}{\partial \dot{\theta}}\right)-\frac{\partial L}{\partial \theta}=M \\
\frac{7}{12} m r^3 l^2 \ddot{\theta}+B \dot{\theta}+k \theta=M \\
\ddot{\theta}+D \dot{\theta}+K \theta=\frac{N}{A} P_{i n}
\end{gathered}
$$
where $A=\frac{7}{12} m \gamma^3 l^2$, the damping coefficient $D=\frac{B}{A}$, and the stiffness coefficient $K=\frac{k}{A}$.

$N$ is the product cross-sectional area of the soft finger and the distance between the center of area and the center of rotation. Fig. \ref{fig2}e shows a cross-section of the actuator where the shaded area is the hollow chamber of the inner layer of the soft finger. Defining $e$ as the total width of the finger, $a$ as the width of the inner wall of the soft finger, $h$ as the total length of the soft finger and $l_{\text {arm }}$ as the distance from the fixed end of the beam-shaped finger to where the force is exerted, we can express
$$
N=\int_{a+l_{\text {arm }}}^{a+b+l_{\text {arm }}}(e-2 a) \cdot h \cdot d h .
$$

We can now define the transfer function of our soft finger as:
$$
\frac{\theta(s)}{P_{\text {in }}(s)}=\frac{\frac{N}{\Lambda}}{s^2+D s+K}
$$
\subsection{Experimental Setup of Spring and Damping Parameters Using Logarithmic Decrement Technique }
To capture the damped harmonic oscillation for the stiffness and damping parameter acquisition, we mount one end of the soft finger on an aluminum extrusion and affix it horizontally (SFig. 2). We then push the tip of the soft finger from its resting position and release it from an offset position. This causes the soft finger to exhibit a harmonic bending behavior as it returns to its resting position (Fig. \ref{fig2}g). The real-time angular position of the soft finger is then recorded using the embedded flex sensor. The flex sensor is interfaced with a National Instruments myRIO-1900, which performs data acquisition at a rate of 100 Hz. We perform this 10 times for each finger to acquire average values of the stiffness and damping coefficients (Supplementary Table 1).

\subsection{Position and Force Output Modeling Experimental Setup }
To measure the experimental angular position as a function of the pressure input, we set up the soft finger similar to how we set up to find the harmonic oscillation behavior in the Methods section Experimental Setup of Spring and Damping Parameters using Logarithmic Decrement Technique. We mount one end of the soft finger on an aluminum extrusion and use the embedded flex sensor’s data to acquire the angular position. We then load the air pressure from 0 to 200 kPa to actuate the soft finger.

To measure the experimental force output as a function of the pressure input, we mount one end of the soft finger on an aluminum extrusion and measure the force output at the other end using an Omega Force Gauge (DFG55-50) as shown in (SFig. 3). We then load air pressure to actuate the soft finger. The force measurements are only taken when the soft finger is bent to a $90\degree$ angle. Therefore, each finger required different pressure inputs to start the force measurement at a $90\degree$ angle.  

\subsection{Finite Element Analysis of Soft Fingers }
We perform Finite Element Analysis (FEA) for each soft finger using Ansys Workbench 2015. We set the material of the soft finger as ACEO silicone shore A30, and used the first-order Ogden model as the constitutive model. The parameters $\mu_1=75499 Pa$ and $a_1=5.836$ were set for the Ogden model, which are obtained by performing tensile test experiments on the silicone rubber material (SFig. 1). For the ring structure, we set the material to standard PLA with a Young’s modulus of 1.1 GPa and a Poisson’s ratio of 0.42. In the FEA simulation, we apply 0-200 kPa air pressure to the inner wall of the index, ring, and little fingers; 0-150 kPa for the middle; and 0-250 kPa for the thumb. These ranges bend the respective fingers from $0\degree$ to $180\degree$ (Fig. \ref{fig3}g and SFig. 5).  

\section{Discussion }
We present the PRBM+LDM method as a convenient and straightforward estimation technique to model a beam-shaped soft actuator's behavior. PRBM+LDM does not rely on the human operator's experimental experience. The amount of time it takes to acquire the damped oscillation behavior of the soft actuator is shorter than that of rigorous tensile tests. PRBM+LDM does not require altering or disassembling the soft actuator's design to acquire the required parameters of the model. Therefore, the dynamic equation acquired using the PRBM and the stiffness and damping coefficients acquired through the LDM experiment closely model the actual behavior of the assembled soft actuator. This leads to reduced error in estimating position and force output as a function of the pressure input.

In our LDM experiments, we consistently observed asymmetric waveforms (SFig. 4) across all trials. This phenomenon likely stems from the unique asymmetric design of the finger's bellow structures. Specifically, the bellows in the dorsal area are more pronounced than those in the ventral area (Fig. \ref{fig4}). Moreover, the soft finger is engineered to bend exclusively in one direction. Despite these asymmetries, the asymmetric waveforms should have little impact on the parameter estimation quality since we only measure the peaks for the LDM in the direction where the soft finger bends. 

We compared the accuracy between the PRBM+LDM dynamic equation and FEA in modeling the angular position of the soft finger as a function of pressure input. The experimental results aligned better with the PRBM+LDM results. This posits the ability of PRBM+LDM to describe the kinematics of soft actuators more accurately than FEA. A relevant advantage of PRBM+LDM compared to FEA is that it does not rely on material properties as parameters; instead, it relies on behavioral parameters that take into account the full assembly of the actuator. This reduces the error between the modeling and the experimental values since the fabrication and assembly of the actuator influence its final behavior. 

We also inspected the accuracy of the PRBM+LDM dynamic equation to model the output force of the soft finger as a function of pressure input. We found an average of $0.375N$ discrepancy between the real output force and the modeled force and a maximum difference of $0.53N$. The discrepancy increases as the force output increases. This increasing discrepancy may be caused by slipping between the tip of the soft actuator and the probe of the force sensor. A way to alleviate the error is to redesign the end of the soft finger that comes in contact with its environment, potentially improving measurement accuracy.

Conventionally, in evaluating soft robot hand systems, grasping tasks are usually carried out. However, in our work, we performed pinching tasks for small objects. For a dexterous manipulation study, in general, there are three categories of tasks: pick and place, in-hand manipulation, and relative motions between two pieces of one object (e.g., using one hand to open/close a case). Among these, multi-fingered pick-and-place tasks are the easiest to achieve, as the involvement of more than two fingers increases stability in securing the object. In contrast, pinching tasks, which use only two fingers, provide less support and are, therefore, more challenging, particularly with small objects. For this reason, we focused on pinching tasks to explore more demanding control scenarios. Tackling in-hand manipulation and relative motion between two pieces of one object are also worth tackling. However, to achieve this, the soft robot must be able to move side-to-side or an abduction-adduction movement on top of the already achieved flexion-extension movement. Currently, our system does not allow for these sophisticated movements.

We applied the PRBM+LDM as a force controller for the soft robot hand. However, the soft robot hand does not include a force sensor on its surface due to the challenge of fabricating reliable soft sensors. Therefore, we incorporated a force estimator as a feedback to our force controller. The force controller enabled the soft robot hand to pinch small objects, i.e., a brass coin, a screwdriver, and a potato chip. There was a noticeable difference between the estimated output force and the experimental output force when we inspected the force modeling of the PRBM+LDM's dynamic model, especially at higher pressure input (Fig. \ref{fig3}b). However, compared to the simple constant pressure grasping control, the PRBM+LDM-based had a higher success rate in the pinching task, especially for the brass coin. Handling the brass coin at its edge may have proven more feasible for the PRBM+LDM closed-loop force controller since such a task requires more accurate force control. This successful handling showcases the functionality of the PRBM+LDM-based force controller, especially since traditionally soft actuators perform grasping tasks through blind grasping, where the input air pressure is tuned manually to a specific value and applied to a wide range of objects with different physical properties \cite{Puhlmann2022}. Some also tune the pressure input value depending on the object type \cite{luo2022}. This direct pressure control method is easy to practice. However, it has limitations, such as not generating accurate forces on the objects, which could decrease the success rate of pinching and show difficulty in holding the objects at thin contact areas (e.g., the edge of the brass coin). Furthermore, manually tuning the pressure input on each soft finger is unintuitive and time-consuming since it is challenging for the human operator to estimate the right amount of pinching force by directly tuning the input pressure for different objects. Therefore, using a force controller is much more conducive for soft robot applications.

It is important to note that the human operator holding the object plays an essential role in the success of the pinching tasks. The human operator needed to orient the object's desired contact points along the median plane of the finger pair for successful pinching. This careful orientation was especially important for the brass coin when the human operator needed to make sure the symmetric plane of the brass coin was as coincident as possible with the median plane of the finger pair. There is, however, flexibility in the positioning of the object as long as it is along the median plane and the finger pair can reach the object. We aim to embed soft tactile force sensors, which measure the contact force on the soft fingers, to improve the PRBM+LDM-based force controller. The soft sensors will provide actual force feedback in our control loop to improve the model of the force controller and the functional task success rate of the soft robot hand.

From Table \ref{tab3} and Table \ref{tab4}, the success of the pinching was higher when the finger pair was composed of the thumb and the little finger. This higher success rate may be attributed to the short length of the little finger, which promotes a decrease in misalignment of the finger since forces at its tip generate less momentum. Therefore, in future works, iteration of the design of the soft fingers to have more constraints for horizontal movement will be considered to increase the success rate of pinching for the other three finger pairs.

As a physical HRI application, we plan to use the soft robot hand as a wearable soft hand exoskeleton for rehabilitation. The position and force controllers we developed serve as preliminary controllers as we will develop in our future works a position controller that can properly aid hand movement trajectory of the user to perform proper grasping movements and a force controller that will employ resistive and assistive guiding forces for the user to perform hand rehabilitation exercises. This wearable application also connotes the importance of real-time model-based control. Therefore, extending the PRBM+LDM method by incorporating the human-joint dynamic model will be part of our future works to compensate for the human joint stiffness and damping. We will then execute a pilot test to evaluate the performance of the extended controller with the soft robot hand attached to an anthropomorphic dummy hand. After such evaluation, a pilot test on actual human subjects will then be feasible.

Furthermore, we will also develop an impedance controller to control position and force exertion at the same time. The impedance controller will allow us to provide trajectory guidance and apply the assistive or resistive force needed to generate virtual force environments essential in rehabilitative applications.

Overall, the PRBM+LDM method presents a streamlined and effective approach for modeling the dynamics of soft actuators. It offers a straightforward process that overcomes computationally inefficient methods and complex experimental setups. It paves the way for convenient solutions for soft robotic modeling in the applications of physical HRI.

\section{Acknowledgment}
The team would like to express our deepest appreciation to Parth Patki for his exceptional graphic rendering work. We also extend our gratitude to Ashwin Hingwe for his meticulous mechanical design around the thumb. Their dedication and contributions have been invaluable to our project’s success.
\bibliographystyle{MSP}
\bibliography{sn-bibliography}

\begin{thebibliography}{10}
\providecommand{\url}[1]{\texttt{#1}}
\providecommand{\urlprefix}{URL }

\bibitem{Das2019}
A.~Das, M.~Nabi,
\newblock In \emph{2019 International Conference on Computing, Communication, and Intelligent Systems (ICCCIS)}. IEEE, \textbf{2019} 306--311.

\bibitem{Kim2013}
Y.-J. Kim, S.~Cheng, S.~Kim, K.~Iagnemma,
\newblock \emph{IEEE Transactions on Robotics} \textbf{2013}, \emph{29}, 4 1031.

\bibitem{Low2014}
J.-H. Low, I.~Delgado-Martinez, C.-H. Yeow,
\newblock \emph{Journal of Medical Devices} \textbf{2014}, \emph{8}, 4 044504.

\bibitem{Liu2016}
Y.~Liu, H.~Xie, H.~Wang, W.~Chen, J.~Wang,
\newblock In \emph{2016 IEEE/SICE International Symposium on System Integration (SII)}. \textbf{2016} 403--408.

\bibitem{Runciman2019}
M.~Runciman, A.~Darzi, G.~P. Mylonas,
\newblock \emph{Soft Robotics} \textbf{2019}, \emph{6}, 4 423, pMID: 30920355.

\bibitem{Schmitt2018}
F.~Schmitt, O.~Piccin, L.~Barb{\'e}, B.~Bayle,
\newblock \emph{Frontiers in Robotics and AI} \textbf{2018}, \emph{5} 84.

\bibitem{Pan2022}
M.~Pan, C.~Yuan, X.~Liang, T.~Dong, T.~Liu, J.~Zhang, J.~Zou, H.~Yang, C.~Bowen,
\newblock \emph{Advanced Intelligent Systems} \textbf{2022}, \emph{4}, 4 2100140.

\bibitem{Rus2015}
D.~Rus, M.~T. Tolley,
\newblock \emph{Nature} \textbf{2015}, \emph{521}, 7553 467.

\bibitem{Thurutel2018}
T.~George~Thuruthel, Y.~Ansari, E.~Falotico, C.~Laschi,
\newblock \emph{Soft robotics} \textbf{2018}, \emph{5}, 2 149.

\bibitem{Shepherd2011}
R.~Shepherd, F.~Ilievski, W.~Choi, S.~Morin, A.~Stokes, A.~Mazzeo, X.~Chen, M.~Wang, G.~Whitesides,
\newblock \emph{Proceedings of the National Academy of Sciences of the United States of America} \textbf{2011}, \emph{108} 20400.

\bibitem{onal2017soft}
C.~D. Onal, X.~Chen, G.~M. Whitesides, D.~Rus,
\newblock In \emph{Robotics Research: The 15th International Symposium ISRR}. Springer, \textbf{2017} 525--540.

\bibitem{martinez2013robotic}
R.~V. Martinez, J.~L. Branch, C.~R. Fish, L.~Jin, R.~F. Shepherd, R.~Nunes, Z.~Suo, G.~M. Whitesides,
\newblock \emph{Advanced materials} \textbf{2013}.

\bibitem{tolley2014resilient}
M.~T. Tolley, R.~F. Shepherd, B.~Mosadegh, K.~C. Galloway, M.~Wehner, M.~Karpelson, R.~J. Wood, G.~M. Whitesides,
\newblock \emph{Soft robotics} \textbf{2014}, \emph{1}, 3 213.

\bibitem{correll2014soft}
N.~Correll, {\c{C}}.~D. {\"O}nal, H.~Liang, E.~Schoenfeld, D.~Rus,
\newblock In \emph{Experimental Robotics: The 12th International Symposium on Experimental Robotics}. Springer, \textbf{2014} 227--240.

\bibitem{Baghat2019}
S.~Bhagat, H.~Banerjee, Z.~T. Ho~Tse, H.~Ren,
\newblock \emph{Robotics} \textbf{2019}, \emph{8}, 1 4.

\bibitem{Iyengar2020}
K.~Iyengar, G.~Dwyer, D.~Stoyanov,
\newblock \emph{International Journal of Computer Assisted Radiology and Surgery} \textbf{2020}, \emph{15}, 7 1157.

\bibitem{Morimoto2021}
R.~Morimoto, S.~Nishikawa, R.~Niiyama, Y.~Kuniyoshi,
\newblock In \emph{2021 IEEE 4th International Conference on Soft Robotics (RoboSoft)}. IEEE, \textbf{2021} 141--148.

\bibitem{Zhao2021}
Q.~Zhao, J.~Lai, K.~Huang, X.~Hu, H.~K. Chu,
\newblock \emph{IEEE/ASME Transactions on Mechatronics} \textbf{2021}, \emph{27}, 5 2511.

\bibitem{Li2022}
Y.~Li, X.~Wang, K.-W. Kwok,
\newblock In \emph{2022 IEEE/RSJ International Conference on Intelligent Robots and Systems (IROS)}. IEEE, \textbf{2022} 7074--7081.

\bibitem{AlAttar2023}
A.~AlAttar, I.~B. Hmida, F.~Renda, P.~Kormushev,
\newblock In \emph{2023 IEEE International Conference on Soft Robotics (RoboSoft)}. \textbf{2023} 1--7.

\bibitem{Almanzor2023}
E.~Almanzor, F.~Ye, J.~Shi, T.~G. Thuruthel, H.~A. Wurdemann, F.~Iida,
\newblock \emph{IEEE Transactions on Robotics} \textbf{2023}.

\bibitem{Santina2023}
C.~Della~Santina, C.~Duriez, D.~Rus,
\newblock \emph{IEEE Control Systems Magazine} \textbf{2023}, \emph{43}, 3 30.

\bibitem{Trivedi2008}
D.~Trivedi, A.~Lotfi, C.~D. Rahn,
\newblock \emph{IEEE Transactions on Robotics} \textbf{2008}, \emph{24}, 4 773.

\bibitem{Robert2010}
R.~J. Webster~III, B.~A. Jones,
\newblock \emph{The International Journal of Robotics Research} \textbf{2010}, \emph{29}, 13 1661.

\bibitem{Camarillo2009}
D.~B. Camarillo, C.~R. Carlson, J.~K. Salisbury,
\newblock In \emph{Experimental Robotics: The Eleventh International Symposium}. Springer, \textbf{2009} 271--280.

\bibitem{Kapadia2011}
A.~Kapadia, I.~D. Walker,
\newblock In \emph{2011 IEEE/RSJ international conference on intelligent robots and systems}. IEEE, \textbf{2011} 1087--1092.

\bibitem{Kapadia2014}
A.~D. Kapadia, K.~E. Fry, I.~D. Walker,
\newblock In \emph{2014 IEEE/RSJ International Conference on Intelligent Robots and Systems}. IEEE, \textbf{2014} 329--335.

\bibitem{Penning2011}
R.~S. Penning, J.~Jung, J.~A. Borgstadt, N.~J. Ferrier, M.~R. Zinn,
\newblock In \emph{2011 IEEE International Conference on Robotics and Automation}. \textbf{2011} 4822--4827.

\bibitem{Penning2012}
R.~S. Penning, J.~Jung, N.~J. Ferrier, M.~R. Zinn,
\newblock In \emph{2012 IEEE International Conference on Robotics and Automation}. IEEE, \textbf{2012} 5392--5397.

\bibitem{Hannan2003}
M.~W. Hannan, I.~D. Walker,
\newblock \emph{Journal of Robotic Systems} \textbf{2003}, \emph{20}, 2 45.

\bibitem{Li2018}
M.~Li, R.~Kang, S.~Geng, E.~Guglielmino,
\newblock \emph{Transactions of the Institute of Measurement and Control} \textbf{2018}, \emph{40}, 11 3263.

\bibitem{Jones2006}
B.~A. Jones, I.~D. Walker,
\newblock \emph{IEEE Transactions on Robotics} \textbf{2006}, \emph{22}, 1 43.

\bibitem{Renda2018}
F.~Renda, F.~Boyer, J.~Dias, L.~Seneviratne,
\newblock \emph{IEEE Transactions on Robotics} \textbf{2018}, \emph{34}, 6 1518.

\bibitem{Xu2022}
X.~Wang, C.~Wang, X.~Wang, D.~Meng, B.~Liang, H.~Xu,
\newblock In \emph{2022 IEEE 18th International Conference on Automation Science and Engineering (CASE)}. \textbf{2022} 1933--1939.

\bibitem{Roshanfar2023}
M.~Roshanfar, J.~Dargahi, A.~Hooshiar,
\newblock \emph{Actuators} \textbf{2024}, \emph{13}, 1.

\bibitem{Bui2021}
P.~D. Bui, J.~A. Schultz,
\newblock \emph{Frontiers in Robotics and AI} \textbf{2021}, \emph{8} 749591.

\bibitem{Schultz2022}
J.~A. Schultz, H.~Sanders, P.~D.~H. Bui, B.~Layer, M.~Killpack,
\newblock In \emph{2022 International Conference on Robotics and Automation (ICRA)}. IEEE, \textbf{2022} 3223--3229.

\bibitem{Connolly2015}
F.~Connolly, P.~Polygerinos, C.~J. Walsh, K.~Bertoldi,
\newblock \emph{Soft Robotics} \textbf{2015}, \emph{2}, 1 26.

\bibitem{Duriez2013}
C.~Duriez,
\newblock In \emph{2013 IEEE international conference on robotics and automation}. IEEE, \textbf{2013} 3982--3987.

\bibitem{Vavourakis2011}
V.~Vavourakis, A.~Kazakidi, D.~P. Tsakiris, J.~A. Ekaterinaris,
\newblock In \emph{COUPLED IV: proceedings of the IV International Conference on Computational Methods for Coupled Problems in Science and Engineering}. CIMNE, \textbf{2011} 147--159.

\bibitem{Bandopadhya2008}
D.~Bandopadhya, B.~Bhattacharya, A.~Dutta,
\newblock \emph{Journal of Intelligent Material Systems and Structures} \textbf{2008}, \emph{20} 51.

\bibitem{Tang2019}
Z.~Q. Tang, H.~L. Heung, K.~Y. Tong, Z.~Li,
\newblock In \emph{2019 International Conference on Robotics and Automation (ICRA)}. \textbf{2019} 4004--4010.

\bibitem{Howell2013}
L.~L. Howell, S.~P. Magleby, B.~M. Olsen, J.~Wiley,
\newblock \emph{Handbook of compliant mechanisms},
\newblock Wiley Online Library, \textbf{2013}.

\bibitem{Jadhav2019}
S.~Jadhav, A.~Kavitake, A.~Lawand \textbf{2019}.

\bibitem{Polygerinos2017}
P.~Polygerinos, N.~Correll, S.~A. Morin, B.~Mosadegh, C.~D. Onal, K.~Petersen, M.~Cianchetti, M.~T. Tolley, R.~F. Shepherd,
\newblock \emph{Advanced Engineering Materials} \textbf{2017}, \emph{19}, 12 1700016.

\bibitem{zou2024}
S.~Zou, S.~Picella, J.~de~Vries, V.~G. Kortman, A.~Sakes, J.~T. Overvelde,
\newblock \emph{Nature Communications} \textbf{2024}, \emph{15}, 1 539.

\bibitem{hargrove2022}
B.~Hargrove, A.~Nastevska, M.~Frecker, J.~Jovanova,
\newblock \emph{Mechanism and Machine Theory} \textbf{2022}, \emph{176} 105017.

\bibitem{Puhlmann2022}
S.~Puhlmann, J.~Harris, O.~Brock,
\newblock \emph{IEEE Transactions on Robotics} \textbf{2022}, \emph{PP} 1.

\bibitem{luo2022}
Y.~Luo, K.~Wu, A.~Spielberg, M.~Foshey, D.~Rus, T.~Palacios, W.~Matusik,
\newblock In \emph{Proceedings of the 2022 CHI Conference on Human Factors in Computing Systems}. \textbf{2022} 1--13.

\end{thebibliography}

\end{document}